\documentclass{article}

\usepackage{arxiv}

\usepackage[utf8]{inputenc} % allow utf-8 input
\usepackage[T1]{fontenc}    % use 8-bit T1 fonts
\usepackage{hyperref}       % hyperlinks
\hypersetup{
    colorlinks=true,
    linkcolor=blue,
    filecolor=magenta,      
    urlcolor=cyan,
    pdftitle={Overleaf Example},
    pdfpagemode=FullScreen,
    }
\usepackage{url}            % simple URL typesetting
\usepackage{booktabs}       % professional-quality tables
\usepackage{amsfonts}       % blackboard math symbols
\usepackage{nicefrac}       % compact symbols for 1/2, etc.
\usepackage{microtype}      % micro typography
\usepackage{lipsum}
\usepackage{caption}
\usepackage{subcaption}
\usepackage{fancyhdr}       % header
\usepackage{graphicx}       % graphics
\graphicspath{{media/}}     % organize your images and other figures under the media/ folder
\usepackage[linesnumbered,ruled,vlined]{algorithm2e} %By DG
\SetKwInput{KwInput}{Input}              % Set the Input
\SetKwInput{KwOutput}{Output} 

\title{Class adaptive threshold and negative class guided noisy annotation robust Facial Expression Recognition}

\author{
  Darshan Gera \\
  SSSIHL, Brindavan Campus \\
  Bengaluru, Karnataka, India\\
  \texttt{darshangera@sssihl.edu.in} \\
  \And
  Badveeti Naveen Siva Kumar\\
  SSSIHL, Prasanthi Nilayam Campus \\
  Sri Sathya Sai District, Andhra Pradesh, India \\
  \texttt{bnaveensivakumar@gmail.com} \\
  \AND
 Bobbili Veerendra Raj Kumar\\
  SSSIHL, Prasanthi Nilayam Campus \\
  Sri Sathya Sai District, Andhra Pradesh, India \\
  \texttt{veerendra.rajkumar@gmail.com} \\
  \And
S Balasubramanian \\
  SSSIHL, Prasanthi Nilayam Campus \\
  Sri Sathya Sai District, Andhra Pradesh, India\\
  \texttt{sbalasubramanian@sssihl.edu.in} \\
}

\begin{document}
\maketitle

\begin{abstract}
The hindering problem in facial expression recognition (FER) is the presence of inaccurate annotations referred to as noisy annotations in the datasets. These noisy annotations are present in the datasets inherently because the labeling is subjective to the annotator, clarity of the image, etc. Recent works use sample selection methods to solve this noisy annotation problem in FER. In our work, we use a dynamic adaptive threshold to separate confident samples from non-confident ones so that our learning won't be hampered due to non-confident samples. Instead of discarding the non-confident samples, we impose consistency in the negative classes of those non-confident samples to guide the model to learn better in the positive class. Since FER datasets usually come with 7 or 8 classes, we can correctly guess a negative class by 85\% probability even by choosing randomly. By learning "which class a sample doesn't belong to", the model can learn "which class it belongs to" in a better manner. We demonstrate proposed framework's effectiveness using quantitative as well as qualitative results. Our method performs better than the baseline by a margin of 4\% to 28\% on RAFDB and 3.3\% to 31.4\% on FERPlus for various levels of synthetic noisy labels in the aforementioned datasets.
% We have shown the effectiveness of this framework based on various experiments. We have used ResNet-18 as a baseline and showed that negative class classification along with positive class classifier helps the model to improve by a margin of 4\% to 28\% improvement over baseline on various levels of symmetric noise of RAFDB and 3.3\% to 31.4\% on various levels of symmetric noise of FERPlus over baseline.
\end{abstract}

% keywords can be removed

\begin{figure}[hbt!]
\centering
  \includegraphics[scale=0.4]{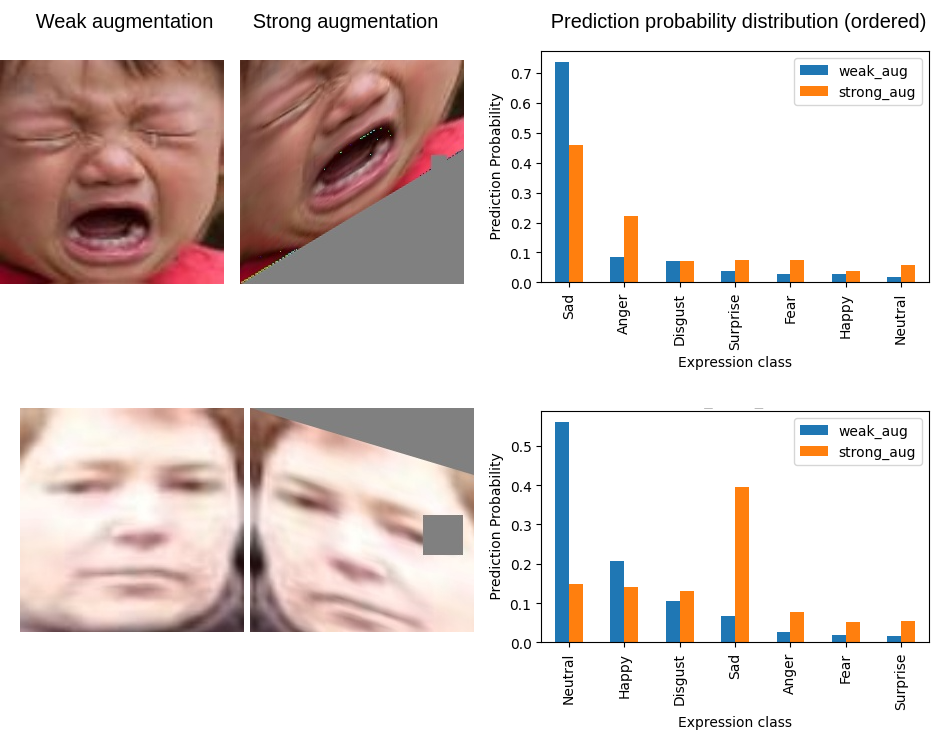}
  \caption{This figure shows a) weakly augmented image, b)strongly augmented view, and c) the prediction probabilities by the ResNet-18 model trained on 10\% synthetic label noise dataset of RAFDB.
  On the top set of images, the model predicted the correct label on both weak and strong augmentations whereas, in the bottom set of images, the model predicted different labels on different augmentations. In spite of not having an overlap in predictions of the most probable class, the model does have an overlap in dissimilarity (least probable class) in both cases.}
  % [Give a), b), c) and d) subheadings under columns after adding input image, remove image name in the figure]}
  \label{fig:motiv}
\end{figure}

\section{Introduction}
In recent years, significant progress has been made towards developing deep learning (DL) based robust facial expression recognition (FER) systems \cite{oadn, scan, LNLAttenNet, ran, FDRL}. Making machines understand human emotions and intentions through the use of facial features is the goal of automatic facial expression recognition. Numerous real-world applications of FER exist, including the detection of driver weariness \cite{R1_App5}, mental health analysis \cite{R1_App1}, increasing student-teacher interaction in distance learning environments \cite{R1_App2}, virtual assistants \cite{R1_App3}, social robots \cite{ R1_App4}, and others. In a supervised setting, learning positive class is quite faster but if there are noisy labeled samples(i.e samples with inaccurate labels) in the dataset, then the model starts learning incorrect features leading to poor generalization. We can subjugate this if we train our model only on the confident samples but getting clean data is a very expensive and tiring process. Labels are subjective to the annotator as well as quality of the image. Because of these factors, datasets inherently have noisy labels.

Several methods tried to alleviate the ill effects of noisy labels on the learning process of the DL model, while training, by choosing samples that  are relatively clean. This is done by choosing samples with low loss \cite{co-teaching,co-teaching_plus, jocr, nct} but these methods lose on learning from clean samples with large loss value might be due to difference in pose or illumination,etc. These methods also use multiple networks for training and \cite{co-teaching,co-teaching_plus,jocr} need to know the noise rate present in the dataset beforehand which won't be available in real-world cases. \cite{gera_ambigious_annotations} uses a single threshold for all the classes. This method doesn't need to know the noise rate beforehand. The threshold that is used for selecting clean samples is based on JS-divergence between two different augmentations of same sample.
% [Write a line more about it]
% In \cite{nlnl}, there are four phases of training, first only with negative learning, then selective negative learning, then selective positive learning followed by semi-supervised learning with relabelling based on certain criteria.

Recent methods such as SCN\cite{scn} use weighted cross-entropy loss (CE loss) where the model itself learns the weights. It groups the weights into two, high importance group and low importance group in a 7:3 ratio after sorting the weights. It relabels the samples among the low-importance group based on a threshold. RUL\cite{RUL} uses two branches, one for learning feature vectors and another for uncertainty vectors. It learns uncertainty vectors by comparing two samples of the same batch. DMUE\cite{DMUE} is an ensemble technique that uses multiple auxiliary branches to mine the latent distribution of a sample. EAC\cite{lfa} uses an attention consistency mechanism between the class activation maps of every class and uses the original image and its flipped version. It uses the CE loss function to learn from all the samples but  the CE loss is not robust to the noisy labels.
 
None of the above-mentioned models cater to inter-class size imbalance problems and intra-class difficulty. To address this issue, we use a dynamic adaptive threshold which we calculate at every mini-batch. So, we don't need to have prior knowledge of noise rate and we use only a single network unlike some of the earlier methods. 

We conducted experiments using a ResNet-18 model trained on a 30\% synthetic noise dataset of RAFDB with CE loss. From the experiment, We made an observation that \iffalse in spite of not being correct at predicting the same most probable class for different augmentations,\fi the model is consistent on the low prediction probabilities(least probable classes) for different augmentations but not always on the most probable class( see Fig\ref{fig:motiv}, on the below set of images the model is consistent on only three least probable classes they are Surprise, Fear, and Anger. whereas in the top set of images, the prediction probabilities are consistent over all the expression classes i.e least being neutral, second least being happy,..., and highest being sad). In other words,  even though the amount of similarity in the most probable class is not the same, there is an overlap in dissimilarity. Inspired by this phenomenon, we want to make use of these least probable classes as negative classes (We are given only with ground truth label, all the others are negative labels for us). Since strong augmentation is a complex augmentation technique, it is not as easy for the model to predict the similar probability distribution of labels as that of weak augmentations'. To do so, the model must learn the correct features or memorize both weak augmentation and strong augmentation together along with the label. 

To make use of negative classes, we can use the prediction probabilities of a single classifier to get a positive class and negative classes as well or we can use another classifier that learns the negative classes separately. To see which is a better option, we conducted another experiment with the ResNet-18 model. We used CE loss only on the confident samples obtained after applying a dynamic adaptive threshold and  using consistency loss on the remaining non-confident samples with different augmentations. Based on other observation from Fig\ref{fig:motiv} that the amount of overlap of dissimilarity may differ for different samples (for eg. in Fig\ref{fig:motiv}, on the top set of images overlap of dissimilarity extended from Neutral to Anger but in the bottom set of images, the overlap of dissimilarity is only for 3 expression classes namely Surprise, Fear and Anger). We would wish to get an optimal overlap of dissimilarity of predictions for the majority of the samples. So we masked the least-k classes where k is a hyper-parameter and used only the prediction probabilities of those that are masked in consistency loss. Expression recognition performance in accuracy is shown in table\ref{cce_kl_exp}.
\begin{table}[hbt!]
\caption{Performance evaluation on RAFDB in the presence of different label noise levels when ResNet-18 is trained with confident samples selected based on dynamic adaptive threshold using cross-entropy loss and on non-confident samples using consistency loss. Here k refers to the number of top-k negative classes used for consistency loss. We give performance in x/y format where x refers to maximum accuracy obtained in training and y refers to average performance on last 5 epochs(i.e from 35 to 40).} \centering
% \small
\setlength{\tabcolsep}{2.8pt} % Default value: 6pt
\renewcommand{\arraystretch}{1.3} 
\begin{tabular}{|c|c|c|c|c|c|c|c|}
\hline
-&k=1&k=2&k=3&k=4&k=5&k=6&k=7\\
\hline
RAFDB&83.409/82.483&83.767/83.148&83.833/82.764&83.409/83.018&83.442/82.822&82.007/67.959&82.431/38.631\\
10\%noise&83.246/82.157&83.148/82.464&82.887/82.235&82.953/82.255&83.05/81.89&82.073/64.152&82.203/55.189\\
60\%noise&74.185/69.511&75.977/70.684&74.12/70.241&76.499/70.225&74.315/70.897&75.195/41.03&76.108/23.122\\
\hline
\end{tabular}
\label{cce_kl_exp}
\end{table}
It has been observed that on higher levels of label noise, performance is greater than the performance of the model that learns only from confident samples obtained by using a dynamic adaptive threshold. But on the lower level noises, there is a performance drop by 5\% on RAFDB and around 4\% on the 10\% synthetic label noise dataset of RAFDB. This shows us the learning of negative labels for non-confident is interfering with the learning of positive class when we use only prediction probabilities from a single classifier. Using two different classifiers, one that predicts positive class and another that predicts negative class proved to be a better option. We show the effectiveness of this approach in section\ref{sec:exp}.

Overall our contributions can be summarized as follow :
\begin{itemize}
    \item We use a single network model unlike previous methods,  ensuring that our model is not computationally expensive.
    \item We deal with inter-class similarities and intra-class difficulties by using dynamic adaptive threshold \cite{DNFER}.
    \item We utilize all the samples in learning the expression features for some it learns what class it is directly from the positive class classifier and for some, it learns what it is not from the negative class classifier which in turn can help in learning what it is eventually.
    \item Our method is an end-to-end- framework that achieves superior performance when the label noise rate is very high in the dataset. In addition, it is also backbone independent.
\end{itemize}

 We showed the effectiveness of  our model on synthetic noisy label datasets generated from  RAFDB\cite{rafdba} and FERPlus\cite{ferplus}, real-world noisy dataset (automatically annotated subset of AffectNet with 0.459M images)\cite{affectnet}. Compared to the baseline model, our method improves the performance in the range of 4\% to 28\%  on RAFDB and in the range of 3.3\% to 31.4\% improvement on FERPlus for various synthetic noisy label datasets generated from RAFDB and FERPlus respectively. Compared to the model that learns using only confident samples obtained by using a dynamic adaptive threshold, our model performs better by a range of 0.03 to 11.7\% on RAFDB and in the range of 0.01 to 1.63\% on FERPlus for various synthetic noisy label datasets generated from RAFDB and FERPlus respectively. 
 
 The remainder of this paper is organized in the following manner: Section \ref{sec:relaed work} gives an overview of other related methods in this field. Section \ref{sec:proposed_method} describes the proposed method and provides the architecture and algorithm for training our framework. Section \ref{sec:exp} shows the effectiveness of the framework using quantitative and qualitative results. We conclude the paper in section \ref{sec:conclusion}.
\section{Related Work}\label{sec:relaed work}
\subsection{General Noisy Label Problem}
\begin{figure}[hbt!]
\centering
  \includegraphics[scale=0.6]{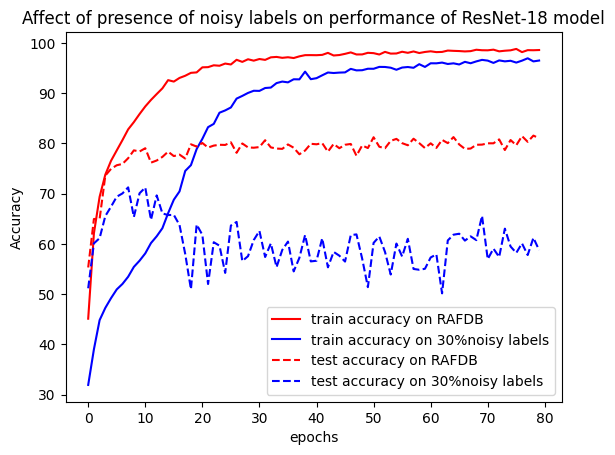}
  \caption{Performance of ResNet-18 trained on RAFDB in the presence of 30\% synthetic noise on RAFDB.}
  \label{fig:memorization}
\end{figure}
% \textbf{INCLUDE THE IMAGE OF RAFDB30NOISE VS RAFDB 0 NOISE}.
If we train a DL model in the presence of noisy labels, due to the strong memorizing capacity of the DL models, the performance of the model hampers as shown in Fig. \ref{fig:memorization}. Several ways are explored in order to solve the noisy label problem. These works can be grouped into few categories such as sample re-weighting, label cleansing, roust loss functions, and selecting clean based on small loss. \cite{mentornet, co-teaching, co-teaching_plus,nct} uses small loss samples to train but they need the prior knowledge of noise rate in the dataset except for \cite{nct}, getting to know the noise rate beforehand is not always possible in real-world scenarios. They also use multiple networks using peer/joint training either with agreement \cite{co-teaching}on the samples selected or by disagreement\cite{co-teaching_plus}.
\cite{nlnl} uses a negative class to deal with noisy labels but it has multiple phases including a semi-supervised learning phase and the loss function in this paper is based on the inversion of prediction probabilities from classifier with a negative label (chosen among the classes other than ground truth label). But when we are training for fewer epochs like 40 (which we do in all our experiments), selective negative learning will lead to prediction probabilities of all classes to be greater than 1/(number of classes) leading to no class being greater than 0.5 when it comes to selective positive learning phase.
Our method works based on sample selection but we don't need to know the noise rate beforehand like the aforementioned methods. We use a dynamic adaptive threshold \cite{DNFER} generated from posterior prediction probabilities from a positive classifier.
\subsection{Noisy Label Problem in FER}
Due to the availability of benchmark datasets like RAFDB\cite{rafdba}, FERPlus\cite{ferplus}, AffectNet\cite{affectnet}, etc. FER has become a well-explored field of research. In spite of this, obtaining good performance when trained on the noisy annotated FER dataset is not a trivial task. Recent works like SCN \cite{scn}, DMUE \cite{DMUE}, IPA2LT \cite{ipa2lt}, CCT \cite{gera_cct1}, and RUL \cite{RUL} have attempted to handle the noisy labels in FER datasets. IPA2LT \cite{ipa2lt} actually deals with inconsistent labels for an image, the real label is learned by maximizing the log-likelihood of multiple inconsistent human-made annotations and machine-predicted annotations. But for this model, we need to have multiple annotations from different annotators. SCN\cite{scn} uses weighted CE loss where weights are learned by the model itself. It tries to segregate the top 70\% and below 30\% of the learned weights as a high-importance group and low-importance group and imposes a loss function such that the model separates these groups by a margin. Among these low-importance groups, it relabels some of the samples based on a criterion. This model suffers from self-confirmation bias and since datasets can have any amount of noise rate in them, separating the learned weights in a $7:3$ ratio is not always optimal. 

RUL\cite{RUL} model has two branches to the backbone model where one branch gives the feature vector and another branch gives the uncertainty vector for each image. For a given batch of images using some random permutation, this model performs a mix-up of feature vectors based on the learned uncertainty values and the model imposes an add-up loss function which forces to predict both the labels correctly from the mixed feature vector. They use the CE loss for this add-up loss function which is not that effective in the presence of label noise. DMUE\cite{DMUE} uses one main branch and as many auxiliary branches as the number of classes for a dataset. The main branch learns from all the samples whereas $i_th$ auxiliary branches learn from the samples other than those that are annotated with $i$ as their label. Consistency is maintained between auxiliary branches and the main branch. It uses CE loss on all the auxiliary branches and weighted CE loss on the main branch where weights are learned based on pairwise uncertainties between the samples of the same mini-batch. This model is computationally very expensive. And the architecture while training is wholly dependent on the number of classes present in the dataset. If we were to have a large number of classes, it would blow up the requirement for computations.

EAC\cite{lfa} uses attention consistency on class activation maps for consistency loss. Activation maps for each class are obtained by performing  the multiplication of weights from the fully connected layer that predicts the class label and the feature maps that are extracted from the penultimate layer of the model. It uses an imbalanced framework where it uses CE loss only on the predictions of the original image but not its flipped version so as to ensure that the model doesn't memorize both the original image and its flipped version to reduce the overall loss. To reduce the memorization effect of the model further, it performs random erasing on the original image.
In spite of its simplicity, the weakness of this model is it uses CE loss on all the images. Cross entropy loss drives the predictions toward the ground truth, which can hamper the learning process.

Unlike DMUE, our proposed model is independent of the number of classes and not expensive in computations. Compared to other methods, our model differs in utilizing only confident samples to be fed to CE loss for positive class prediction, where we obtain confident samples from the dynamic adaptive threshold. None of the above-mentioned methods deal with the intra-class sample size imbalances and inter-class difficulties, we address them by using dynamic adaptive threshold, and instead of discarding the non-confident samples, we learn from them as well by imposing consistency on the negative classes. All the above-mentioned methods try to use loss functions that try to make even the non-confident samples learn the positive class (which can be wrong due to noisy annotations). Here we take an indirect approach to use "not this" philosophy and learn "which class the sample doesn't belong to" on the non-confident samples which in turn can help us to learn "which class it belongs to".
 \begin{figure}[hbt!]
\centering
  \includegraphics[scale=0.225]{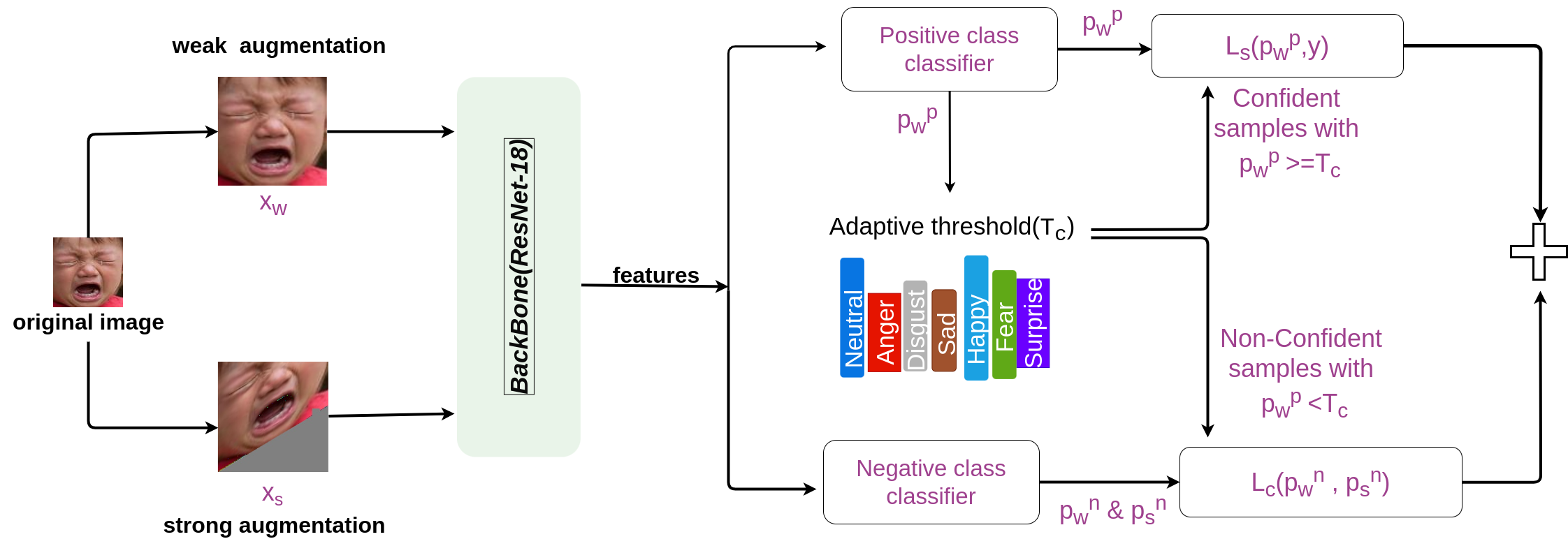}
  \caption{Architecture of proposed model. Here a batch of $x_w and x_s$ (weak and strong augmentations) are sent through the backbone(ResNet-18). Extracted features are sent to the positive class classifier (pcc) as well as negative class classifier(ncc). On the predictions of weak augmentation from pcc, we calculate the adaptive threshold and get confident and non-confident indices for the given batch of samples. Only the confident samples are used for supervised loss$L_s$ and only the non-confident samples are sent to consistency loss $L_c$}
  \label{fig:arch}
\end{figure}

\section{Proposed Method}\label{sec:proposed_method}
 In this section, we first provide the motivation for the proposed method. Then we list out the details of our method. Following it, we provide a training algorithm. The architecture for our model is shown in Fig\ref{fig:arch}
\subsection{Motivation and overview}
Our method is based on the idea of selecting confident samples to learn from in order to ensure that the model doesn't learn incorrect features to predict the labels. Unlike selecting samples having a small loss that leaves out hard samples which can be crucial for the model to better learn features for a given class to generalize better,  
 we use posterior probabilities and generate a threshold to get the confident samples. Datasets do come with multiple challenges like variation in poses, variation in illumination,..., etc.
Due to these reasons even in a given class, there can be easy samples and hard samples because of which the prediction probabilities  which DL model gives may vary. This is the intra-class difficulty.  In order to tackle the inter-class similarities and intra-class difficulties, we use dynamic adaptive threshold \cite{DNFER} to get the confident samples from the positive class classifier. We have conducted an experiment and observed the prediction probability distributions on different augmentations. We have used weak augmentation that includes horizontal flip and center crop. We have used RandAugment\cite{randaugment} for strong augmentation. We observed that the prediction probabilities of the least probable classes are more consistent compared to the most probable class.  Even though the amount of similarity in prediction probabilities of  the most probable class is not the same, there is an overlap in dissimilarity in prediction probabilities of negative classes. Examples of this phenomenon are shown in Fig\ref{fig:motiv}. Even if we predict randomly, we can be correct in guessing the negative label by 85\% given that the datasets that we use are having only 7 or 8 classes. If we ensure the consistency between different augmentations, as training progresses, we can get consistent predictions even in positive classes. Based on an experiment, where we trained the ResNet-18 model with the cross-entropy loss on confident samples obtained using dynamic adaptive threshold and consistency loss on non-confident samples on the negative classes based on the prediction probabilities of the classifier, we have observed that the performance(accuracy) is lower than the performance of the model that uses only confident samples. So consistency loss on negative classes is affecting the learning process of positive classes when we use a single classifier. So we use two classifiers. One is to predict the positive class of the samples where we use a dynamic adaptive threshold. And another negative class classifier that predicts the negative class of the samples. Among these, we apply consistency loss defined in equation\ref{eq:cons} only on the non-confident samples.
% We can see from Fig\ref{fig:arch}Inspired by the effect of negative learning \cite{nlnl} since it is easy to learn what it is not rather than what it is, We use a negative class classifier to guide the learning process.
% 
\subsection{Problem Formulation}
Given a batch of N samples $S =\{(x_{i}, y_{i})\}_{i=1}^{N}$ where each face image $x_{i}$ has an  expression label $y_{i}\in\{1,2,...,C\}$ here C denotes the number of expression classes. The shared backbone network is ResNet-18. It is parameterized by $\theta$. Features from the backbone are classified using two fully connected layers (FC) followed by softmax to obtain prediction probabilities.
The first FC layer is to predict the positive class and the other FC layer predicts the negative class. We use two different augmentations 
 one weak-augmented image and another strongly-augmented image which is denoted by $x_w$ and $x_s$. Prediction probabilities obtained by passing $x_w$ and $x_s$ through the positive class classifier are denoted as $p_w^p$ and $p_s^p$ respectively and through the negative class classifier are denoted as $p_w^n$ and $p_s^n$ respectively. Standard cropping along with horizontal flipping with a probability of 50\% is used for weak augmentations. And for strong augmentation, Randaugment \cite{randaugment} is used. During training, for each mini-batch $S_n$, dynamic adaptive threshold $T_c$ is calculated from the predictions of positive class classifier $p_w^n$ as per equation\ref{eq:dat}, where $X_{c}$ is a set of samples in the current mini-batch with class c.
\begin{equation} \label{eq:dat}
T_{c} = \frac{1}{ X_{c}} \sum_{x \in S_{c}} p^{c}(x;\theta) 
\end{equation}
After finding the dynamic adaptive threshold $T_c$, We choose those images whose ground truth prediction probabilities are greater than $T_c$ for that ground truth label. This can be represented by equation \ref{eq:cleansampleselection} 
\begin{equation} \label{eq:cleansampleselection}
X_{c}^{clean} =  \{ x \in X_{c} \ni p^{c} \geq  T_{c} \}
\end{equation}.
By propagating the losses  only from these confident samples, we learn better from the positive class. Instead of discarding all the other samples from the mini-batch, we use them for feature learning using consistency loss. For this, we take the prediction probabilities from negative class classifier $p_w^n$ and $p_s^n$ and use masked CE loss defined in section\ref{sec:loss-fun}. We say it is masked because the performance of the model dehances if we  were to use prediction probabilities of all classes, we choose top K classes from the negative class classifier predictions and do consistency loss only on those classes. K is a hyper-parameter obtained based on ablation study{\ref{sec:ablation}}.

\subsection{Loss functions}\label{sec:loss-fun}
We use Cross-Entropy loss on a positive class classifier defined by equation\ref{eq:ce}. Here $L_{s}$ represents supervised loss 
\begin{equation}\label{eq:ce}
    L_{s} = \{-\sum_{c=1}^{C} y_{i=1}^{c} \log( p^{c}(x_{i};\theta)\}_{i=1}^{N}
\end{equation}
The consistency loss denoted by $L_c$ on negative class classifier predictions can be obtained using equation\ref{eq:cons}. where $M_k$ represents the mask where $M_k$ becomes zero if the prediction probability for a given class is not among top k else it is one.
\begin{equation}\label{eq:cons}
    L_c  = -\frac{1}{N}\sum_{i=1}^{N}M_kp_w^nlog(p_s^n)
\end{equation}
Overall loss is denoted as $ L_{overall}$\ref{eq:loss}
\begin{equation}\label{eq:loss}
    L_{overall} = L_s+L_c
\end{equation}

%algorithm
\begin{algorithm}
\DontPrintSemicolon
  \KwInput{ Given a model f with parameters $\theta$, dataset $S =\{(x_{i}, y_{i})\}_{i=1}^{N}$, mini-batch size (b), learning rate($\eta$),  number of expression classes (C), total epochs $E_{max}$, warm-up epochs ($E_{warm}$)}
  \KwOut{Updated model parameters $\theta$}
   Initialize $\theta$ randomly.\;
  
   \For{$e = 1,2,..,E_{max}$} 
   {
    Shuffle training samples $\{(x_{i}, y_{i})\}_{i=1}^{N}$\;
    Sample mini-batch $S_{n}$ from S\;
    \For{ each class $c \in \{ 1,2,..,C\}$} 
    {
    \uIf{ $e < E_{warm}$ }{
     Compute loss $L_{s}$ on all samples using Eq. \ref{eq:ce} \;
    }
    \Else{
        Compute dynamic adaptive threshold $T_{c}$ using Eq. \ref{eq:dat}\;
        Select confident samples $S_{c}^{clean}$ from current mini-batch using Eq. \ref{eq:cleansampleselection}\;
    
        Compute supervision loss $L_{s}$ on above selected confident samples using Eq. \ref{eq:ce} \;
        From those non-confident samples, based on negative class classifier prediction probabilities, Compute consistency loss $L_{c}$ using Eq. \ref{eq:cons} \;
        Compute total loss $L$ using Eq. \ref{eq:loss} \;
    }
    }
    %  \uIf{ $e < E_{warm}$ }{
    %  $\alpha = 0$ \;
    % }
  
    % \Else{
    % $\alpha = 0.5$ \;
    % }
    Update model parameters $\theta= \theta - \eta \nabla L_{\theta}$ as per gradient descent rule \;
     }   
   return $\theta$
\caption{Training algorithm}
\label{algo:Training_algorithm}
\end{algorithm}
% Experiments
\section{Experiments}\label{sec:exp}
\subsection{Datasets}
We evaluate our model on three popular real-world benchmark FER datasets RAFDB, FERPlus,  and AffectNet.

\begin{itemize}

   \item \textbf{RAFDB} \cite{rafdba, rafdbb}: The Real-world Affective Face Database (RAFDB) has a basic emotion set of 12271 images for training and 3068 images for testing. Both train and test sets are imbalanced w.r.t sample sizes of different expression classes.
   
   \item 	\textbf{FERPlus} \cite{ferplus}: FERPlus is an extended version of FER2013 \cite{fer2013}. It consists of images with the 8 basic emotions (with contempt), of which 28709 are used for training, 3589 are used for validation and the remaining 3589 for testing.
   
   \item \textbf{AffectNet} \cite{affectnet}: AffectNet is the large dataset with 0.44M manually annotated and 0.459M automatically annotated facial expression images for 8 emotions. We use an Automatically annotated subset of seven classes for training under real noisy conditions and tested on the validation set constituting 3500 images.
   
   \item \textbf{Synthetic noisy annotated datasets}: We randomly change 10\%, 30\%, 50\%, 60\%, 70\%, 80\% labels of training images from RAFDB, FERPlus  to create synthetic noisy annotated datasets. The performance of our model is reported on the corresponding clean test/validation sets.
   
\end{itemize} 
\subsection{Implementation Details}
In the experiments below, we used MTCNN \cite{mtcnn} to recognize and resize the images of facial expressions to 224x224. The PyTorch DL toolbox is used to build our technique, and a single Tesla K40C GPU with 11.4GB RAM is used to run our experiments. The backbone network utilized is ResNet-18, which was previously trained on the MS-Celeb-1M\cite{msceleb} face recognition dataset. In addition to random cropping with 4 pixels and resizing to 224x224, random horizontal flipping with a chance of 0.5 is employed for weak augmentation. RandAugment \cite{randaugment} is used for strong augmentation. From a selection of transformations including contrast adjustment, rotation, color inversion, translation, etc. RandAugment chooses two augmentations at random. Similar to \cite{ran, scn, oadn}, oversampling is used to solve class imbalance issues in AffectNet. The batch size used for training is 128. The model is optimized using the Adam optimizer, with a learning rate of 0.0001 for the backbone and 0.001 for the positive class classifier and negative class classifier (FC layers). We have run the model for 40 epochs.

\subsection{Experiment Results on Synthetic Noisy Annotated Datasets}
In all the experiment results below, we have given performance in accuracy as x/y, where x is the maximum accuracy obtained during the process of training in 40 epochs. y represents the average accuracies of the last 5 epochs (i.e from 36 to 40).
% table
\begin{table}[hbt!]
\setlength{\tabcolsep}{10pt} % Default value: 6pt
\renewcommand{\arraystretch}{1.5} % Default value: 1
\caption{\scriptsize{Performance evaluation comparison on synthetic symmetric label noise on FERPlus}}
    \centering
    \small
    \setlength{\tabcolsep}{3pt}
    \begin{tabular}{c|c|c|c|c|c|c|c}
        \hline
        % \textbf{FERPlus}             &\textbf{SCN}             &\textbf{RUL}             &\textbf{EAC}             &\text
        \textbf{FERPlus}             &10\%noise         &20\%noise         &30\%noise         &50\%noise        &60\%noise          &70\%noise         &80\%noise\\
        \hline
        SCN \cite{scn}                         &84.28             &84.99             &82.47             &75.33             &68.06             &39.43             &37.62\\
        RUL\cite{RUL}                         &85.94             &84.99             &82.75             &77.18             &73.54             &64.07             &43.39\\
        EAC\cite{lfa}                          &\textbf{87.03}             &\textbf{86.07}            &\textbf{85.44}             &81.48	         &79.82	            &74.98	           &62.19\\
        \hline
        NCCTFER                      &86.29    &85.66   &84.79    &\textbf{81.73}    &\textbf{80.20}    &\textbf{75.17}    &\textbf{68.03}\\
        \hline
% \hline
% baseline&85.52/82.689&83.10/70.776&79.18/52.15&77.11/44.86&71.82/32.448&58.9/21.1575 \\
% baseline+pc&86.23/85.298&84.5/83.82&80.42/77.108&79.9489/73.5734&74.91/63.294&59.74/40.868 \\
% % k=5&cce+(negative_classifier)&86.16/85.11&84.6/83.764&80.77/80.36/77.49&78.93/74.26&71.88/63.638&57.02/52.498 \\
% baseline+pc+nc&86.0376/85.5594&84.348/83.832&81.670/77.679&79.63/74.8103&75.10/67.389&68.919/42.55 \\
% %improvement&1.28&$-0.1095$&1.6105&1.2543&0.438&1.6338 \\
% \hline
% improvement&\textbf{0.2614}&\textbf{0.012}&\textbf{0.567}&\textbf{1.2403}&\textbf{4.537}&\textbf{1.688}\\
% improvement&cce+consistency vs cce&-0.24\%&-0.06\%&0.38\%&0.70\%&0.79\%&1.50\% \\

\end{tabular}

\label{syn_fer}
\end{table}

\begin{table}[hbt!]
\setlength{\tabcolsep}{10pt} % Default value: 6pt
\renewcommand{\arraystretch}{1.5} % Default value: 1
\caption{\scriptsize{Performance evaluation comparison on synthetic symmetric label noise on RAFDB}}
    \centering
    \small
    \setlength{\tabcolsep}{3pt}
    %\resizebox{0.5\textwidth}{!}{
    \begin{tabular}{c|c|c|c|c|c|c|c}
    
        \hline
        \textbf{RAFDB}          &10\%noise       &20\%noise    &30\%noise  &50\%noise    &60\%noise    &70\%noise   &80\%noise\\
        \hline
        SCN\cite{scn}                     &82.18           &79.79        &77.46      &73.5         &59.55        &41.98       &38.82\\
        RUL\cite{RUL}                     &86.22           &83.35        &82.06       &73.5        &69.62        &57.66       &36.34\\
        EAC\cite{lfa}                     &\textbf{88.02}  &86.05        &84.42       &80.54	     &76.37	       &68.9	    &47.46\\
        \hline
        NCCTFER                &86.7     &\textbf{86.147} &\textbf{85.169}	&\textbf{81.486} &\textbf{79.73} &\textbf{71.9}	&\textbf{48.89}\\
        \hline
% \hline
% \textbf{RAFDB}&10\%noise&30\%noise&50\%noise&60\%noise&70\%noise&80\%noise\\
% \hline
% baseline&84.7/82&81.2/68&77.1/50&69.2/40&60.26/30.1&41.59/19.774\\
% baseline+\cite{nlnl}loss functions&84.2/82.294&80.54/69.508&75.488/50.16 &71.35/39.64&62.9/29.65&39/18.868\\
% baseline+pc&86.7/86.1656&83.3/80.624&78.94/73.356&75.4/63.824&61.3/47.648&33.1/26.91\\
% baseline+pc+nc&87.05/86.04&82.26/80.5994&80.11/75.92&76.11/68.522&63.16/50.95&42.37/38.6245\\
% % improvement&$-0.1256$&$-0.0246$&2.564&4.706&3.302&11.714\\
% \hline
% improvement&$-0.1256$&$-0.0246$&\textbf{2.564}&\textbf{4.698}&\textbf{3.302}&\textbf{11.714}\\
% \hline

\end{tabular}

\label{syn_raf}

\end{table}

\begin{figure}[hbt!]
\centering
  \includegraphics[scale=0.3]{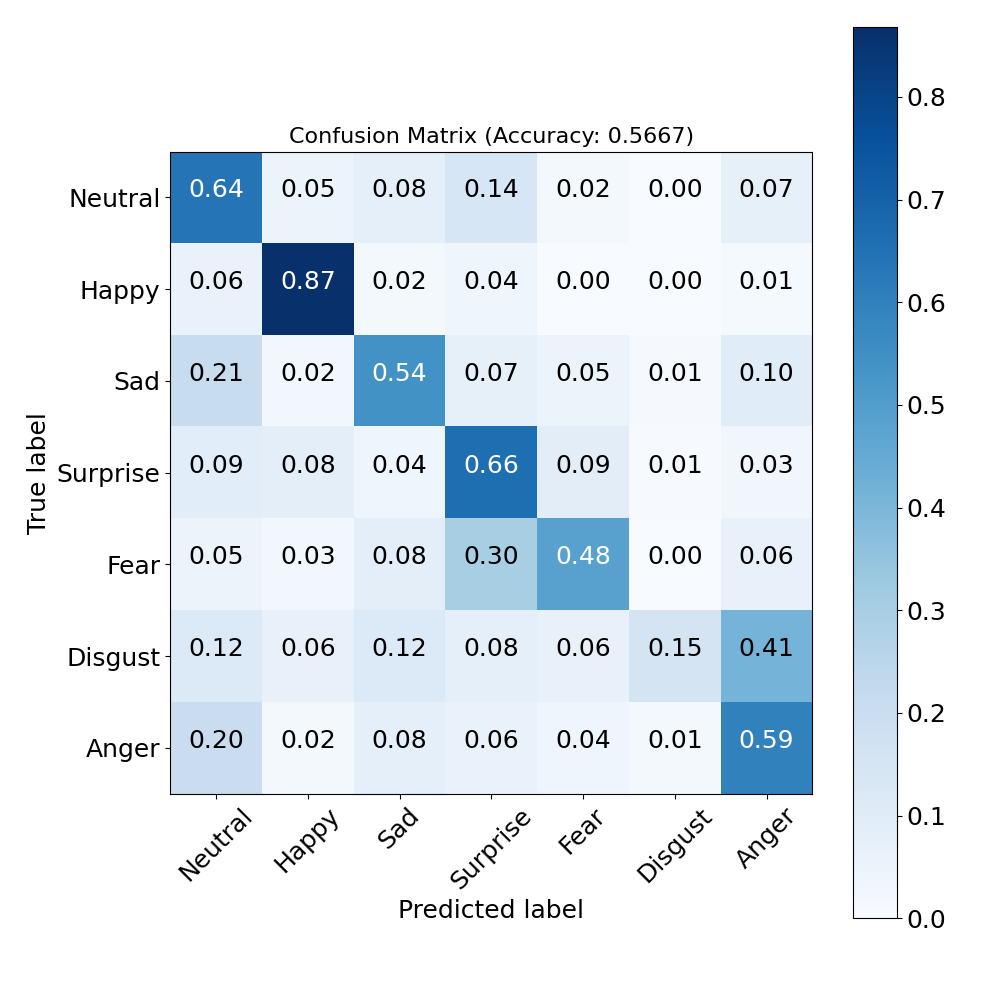}
  \caption{Confusion Matrix of model performance when trained on real noisy data-subset of AffectNet(i.e Automatically Annotated image subset with 0.459M images).}
  \label{fig:cma7}
\end{figure}

Synthetic symmetric noise is manually added on RAFDB, and FERPlus datasets by randomly changing labels in the ratio of 10,30,50,60,70,80\%. We compare our model (referred to as baseline+pc+nc(pc: positive classifier, nc: negative classifier) ) with  ResNet-18 with pre-trained weights of MS-Celeb checkpoint, trained using CE loss (referred to as Baseline)\ref{syn_fer}\ref{syn_raf} and against ResNet-18 with pre-trained weights of MS-Celeb checkpoint, trained using CE loss but only on the confident samples selected using dynamic adaptive threshold(referred to as baseline+pc(positive classifier))\ref{syn_fer}\ref{syn_raf}. Automatically Annotated subset of AffectNet-7 is a challenging noisy dataset because of heavy class imbalance as well as intra-class difficulty and the annotations are generated automatically without human intervention. No method is able to achieve good performance. Our method gave 56.66\% accuracy when tested on the validation set of the AffectNet dataset. The confusion plot for this model is shown in Fig.\ref{fig:cma7}

\begin{figure}
     \centering
     \begin{subfigure}[b]{0.49\textwidth}
         \centering
         \includegraphics[width=\textwidth]{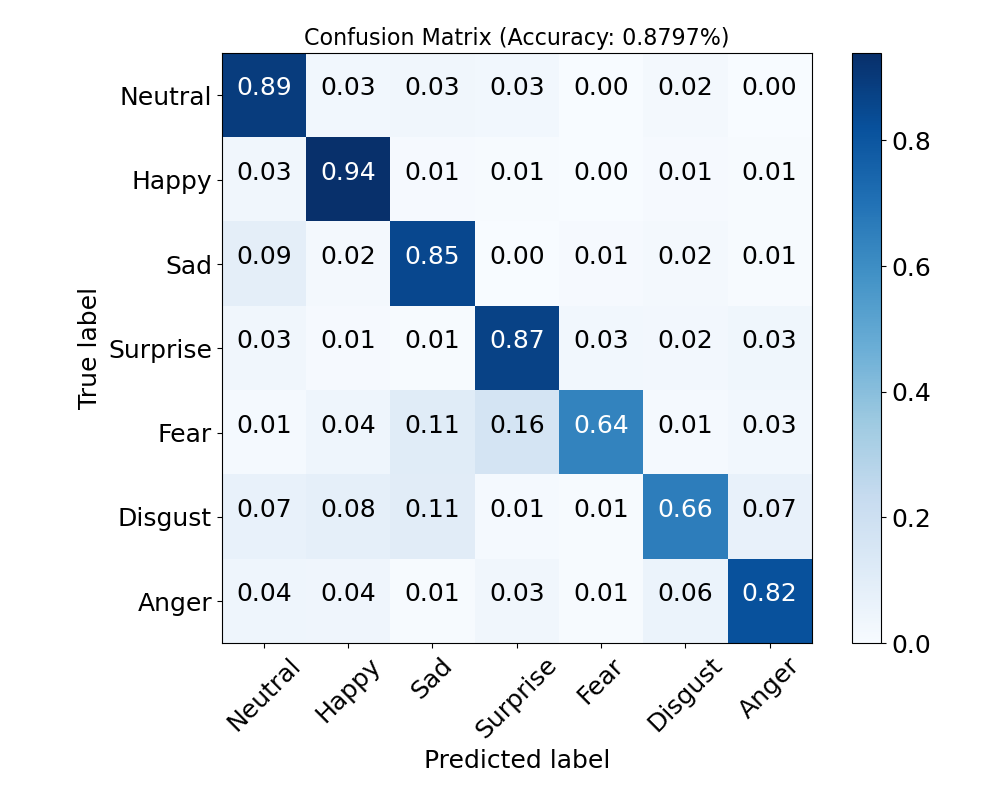}
         % \caption{Confusion Matrix of model performance when trained on RAFDB dataset}
         \label{fig:raf_cm}
     \end{subfigure}
     \hfill
     \begin{subfigure}[b]{0.49\textwidth}
         \centering
         \includegraphics[width=\textwidth]{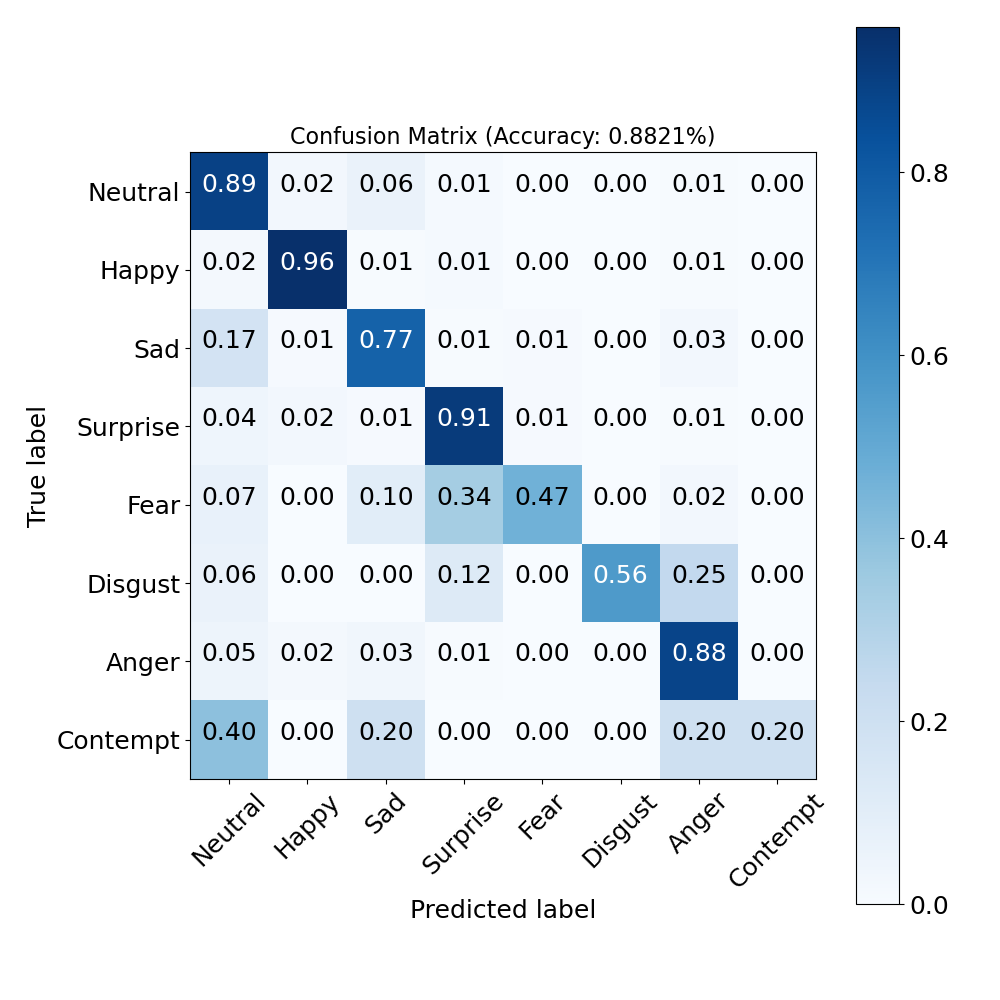}
         % \caption{Confusion Matrix of model performance when trained on FERPlus dataset}
         \label{fig:ferplus_cm}
     \end{subfigure}
     % \hfill
     % \begin{subfigure}[b]{0.3\textwidth}
     %     \centering
     %     \includegraphics[width=\textwidth]{images/Confusion_Matrix_affectnet7_auto_annotated.png}
     %    \caption{Confusion Matrix of model performance when trained on real noisy data-subset of AffectNet(i.e Automatically Annotated image subset with 0.459M images).}
     %    \label{fig:cma7}
        
     % \end{subfigure}
     \caption{Confusion Matrix of model performance when trained on RAFDB(left) and FERPlus(right) respectively}
    \label{fig:cm}
\end{figure}

% \textbf{GENERATE CONFUSION PLOT CHECKPOINT IN .32 mar/aff/all}
\subsubsection{Performance on asymmetric noise}
Apart from symmetric noise, the effectiveness of the model is shown on synthetic asymmetric noise on the RAFDB dataset. For a given expression in RAFDB, we have replaced the expression with its most confused pair in the required percentage. The most confused pairs in RAFDB, 
 based on confusion plots of some SOTA methods are Surprise-Anger, Fear-Surprise, Disgust-Anger, Happy-Neutral, Sad-Neutral, Anger-Happy, and Neutral-Sad. From these confusion pairs, if the former is the ground truth label, we replaced it with the latter for the required amount of noise rate. We have conducted experiments with 10\% to 50\% of asymmetric noise rates as per the above-mentioned method. Our method improves from 1.076 to 7.21\% over the  baseline. The results are shown in the table\ref{asym_raf}.

\begin{table}[hbt!]
\setlength{\tabcolsep}{13pt} % Default value: 6pt
\renewcommand{\arraystretch}{1.5} % Default value: 1
\caption{Performance evaluation in the presence of synthetic asymmetric label noise on RAFDB} 
\centering
\small
\setlength{\tabcolsep}{3pt}
%\resizebox{0.5\textwidth}{!}{
\begin{tabular}{c|c|c|c|c|c}
\hline
RAFDB&10\%noise&20\%noise&30\%noise&40\%noise&50\%noise\\
\hline
Baseline&84.81&80.73&77.28&68.87&48.10\\
% baseline+pc&86.99/85.54&85.52/82.23&83.96/76.36&76.89/60.91&60.92/44.95\\
NCCTFER&85.886& 84.713& 81.799& 75.912&55.31\\
% 86.21/84.94&85.56/82.25&83.73/76.85&78.81/61.27&63.39/48.04\\
\hline
improvement&\textbf{1.076}&\textbf{3.983}&\textbf{4.519}&\textbf{7.042}&\textbf{7.21}\\
\hline
\end{tabular}

\label{asym_raf}
\end{table}

% that of DMUE on AffectNet-7 in the presence of noisy labels
\subsubsection{Visualizations}
We have visualized our model's performance as a confusion matrix as shown in Fig\ref{fig:cm} and \ref{fig:cma7}.
It can be observed that Happy, Surprise, and Neutral are easy classes to predict mainly because of the availability of more samples with these labels. Whereas Disgust and Fear are most confused among all three datasets Fear is most confused with Surprise and Disgust is confused with neutral in FERPlus and RAFDB but Disgust is most confused with Anger in the automatically annotated subset of AffectNet dataset. Contempt in the FERPlus dataset is most confused with Neutral, Anger, and Sad in the mentioned order. 
\begin{figure}
     \centering
     \begin{subfigure}[b]{0.49\textwidth}
         \centering
         \includegraphics[width=\textwidth]{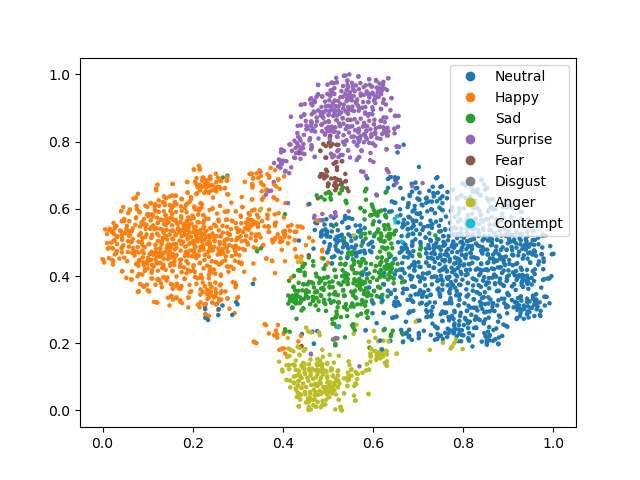}
         \caption{t-sne plot of learned feature vectors from FERPlus dataset}
         \label{fig:ferplus_tsne}
     \end{subfigure}
     \hfill
     \begin{subfigure}[b]{0.49\textwidth}
         \centering
         \includegraphics[width=\textwidth]{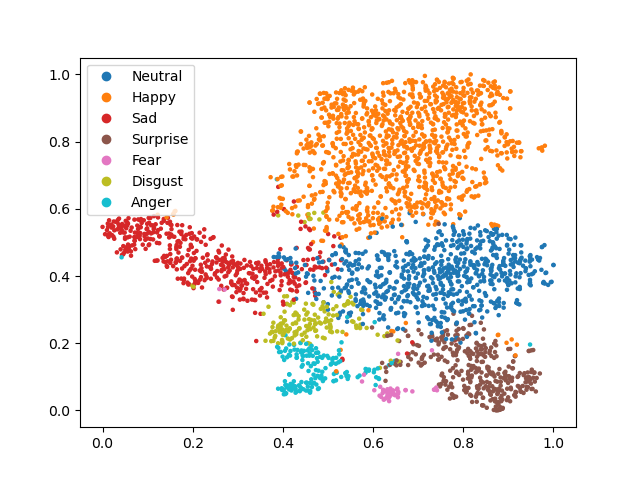}
         \caption{t-sne plot of learned feature vectors from RAFDB dataset}
         \label{fig:raf_tsne}
     \end{subfigure}
     % \caption{Confusion matrices}
    \label{fig:tsne}
\end{figure}
\subsection{Ablation study}\label{sec:ablation}
\begin{figure}[hbt!]
\centering
  \includegraphics[scale=0.45]{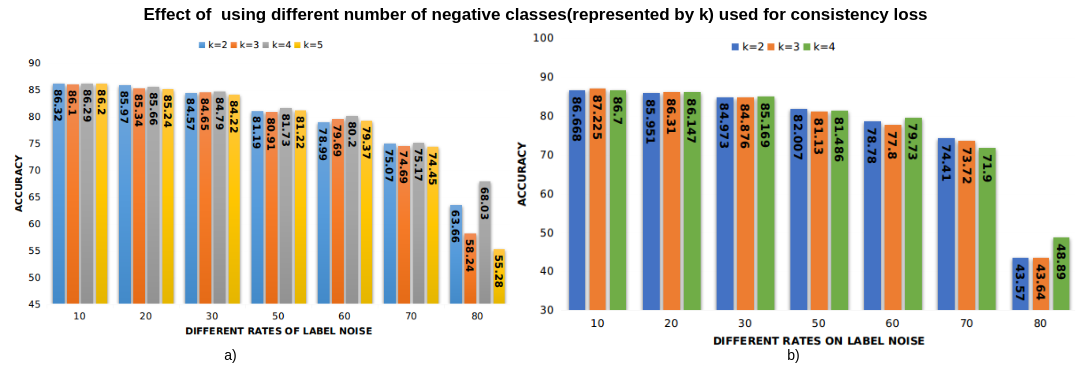}
  \caption{\scriptsize{Effect of k shown on synthetic symmetric label noise of a)FERPlus and b)RAFDB}}
  \label{fig:ablationonk}
\end{figure}

We have seen the effectiveness of masking different number of classes (choosing only top k classes for consistency loss) among the non-confident samples whose prediction probabilities obtained from the positive class classifier failed to be greater than the dynamic adaptive threshold. The hyper-parameter k is used to determine how many classes we are going to use for consistency loss. Results are shown in Fig. \ref{fig:ablationonk}. We have adjusted the value for k as 4 because when we mask the top 4 negative classes, we are achieving good performance over various levels of noisy labels.
Apart from this, we have verified the importance of each component in learning by removing the other components from the model and checking for their performance. Table \ref{components_fer} shows that baseline is far from below compared to the proposed model and effectiveness of learning from negative classes on the non-confident samples is also shown by the increase in performance compared to model that learns only from confident samples. 
\begin{table}[hbt!]

\centering
\setlength{\tabcolsep}{6pt} % Default value: 6pt
\renewcommand{\arraystretch}{1.3} % Default value: 1
\caption{\scriptsize{Ablation study on the importance of learning from positive class and negative class on synthetic symmetric noise on RAFDB (Here Baseline+pc refers to model learning from only confident samples and Baseline+nl refers to baseline with loss functions defined in\cite{nlnl} )}}
\begin{tabular}{|c|c|c|c|c|c|c|}
    \hline
    \textbf{RAFDB}    &10\%noise    &30\%noise    &50\%noise    &60\%noise    &70\%noise    &80\%noise\\
    \hline
    Baseline     &84.7   &81.2    &77.1   &69.2     &60.26     &41.59\\
    Baseline+nl     &84.2    &80.54    &75.488      &71.35     &62.9    &39\\
    Baseline+pc     &86.7     &83.3     &78.94     &75.4     &61.3    &33.1\\
    NCCTFER     &86.7     &85.169     &81.486     &79.73     &71.9    &48.89\\
    
    \hline

\end{tabular}
\label{components_fer}
\end{table}

% \begin{table}[hbt!]

% \centering
% \setlength{\tabcolsep}{6pt} % Default value: 6pt
% \renewcommand{\arraystretch}{1.3} % Default value: 1

% \begin{tabular}{|c|c|c|c|c|c|c|}
% \hline
% \textbf{RAFDB}&10\%noise&30\%noise&50\%noise&60\%noise&70\%noise&80\%noise\\
% \hline
% Baseline&84.7/82&81.2/68&77.1/50&69.2/40&60.26/30.1&41.59/19.774\\
% Baseline+\cite{nlnl}loss functions&84.2/82.294&80.54/69.508&75.488/50.16 &71.35/39.64&62.9/29.65&39/18.868\\
% Baseline+pc&86.7/86.1656&83.3/80.624&78.94/73.356&75.4/63.824&61.3/47.648&33.1/26.91\\
% proposed model&87.05/86.04&82.26/80.5994&80.11/75.92&76.11/68.522&63.16/50.95&42.37/38.6245\\
% % improvement&$-0.1256$&$-0.0246$&2.564&4.706&3.302&11.714\\
% \hline
% improvement&$-0.1256$&$-0.0246$&\textbf{2.564}&\textbf{4.698}&\textbf{3.302}&\textbf{11.714}\\
% \hline

% \end{tabular}
% \caption{Ablation study on importance of learning from positive class and negative class on synthetic symmetric noise on RAFDB (Here Baseline+pc refers to model learning from only confident samples)}
% \label{components_raf}
% \end{table}

\begin{table}[hbt!]
% \begingroup
\centering
\setlength{\tabcolsep}{10pt} % Default value: 6pt
\renewcommand{\arraystretch}{1.5} % Default value: 1
\caption{Performance  on real-world datasets RAFDB and FERPlus (* represents trained on AffectNet and RAFDB combined.)}
% \centering
% \small
% \setlength{\tabcolsep}{3pt}
%\resizebox{0.5\textwidth}{!}{
\begin{tabular}{|c|c|c|}
\hline
Models$\backslash$ Datasets&RAFDB&FERPlus\\
\hline
% baseline&88.2/87.492&87.95/87.242\\

IPA2LT\cite{ipa2lt}&86.77*&-\\
RAN\cite{ran}&86.90&88.55\\
SCN\cite{scn}&88.14*&88.01\\
DMUE\cite{DMUE}&88.76&88.64\\
RUL\cite{RUL}&88.98&88.75\\
% Baseline+pc&88.07&87.72\\
NCCTFER&87.97&88.21\\
\hline
% improvement&0.0394&0.189\\
% \hline

\end{tabular}
%}
% \endgroup
\label{fer_raf}
\end{table}
\subsubsection{Attention Maps}
\begin{figure}[hbt!]
\centering
  \includegraphics[scale=0.45]{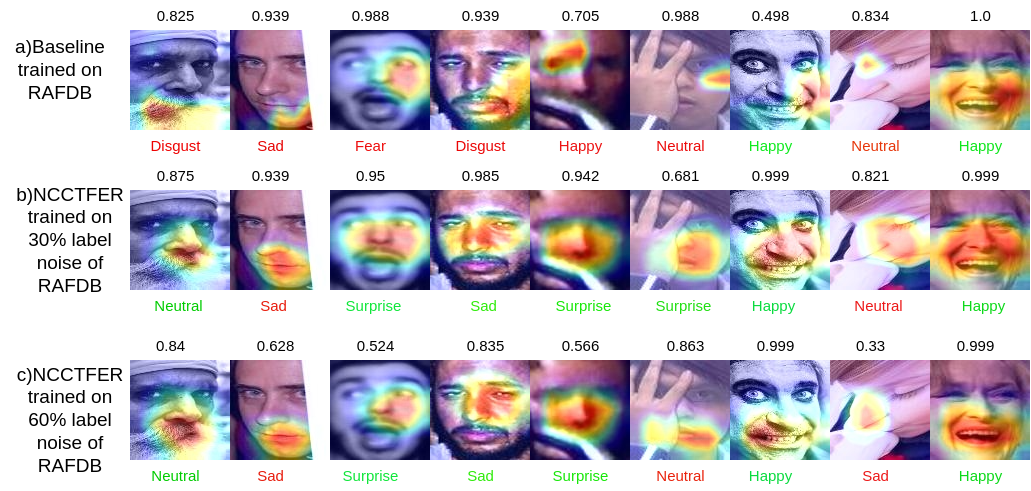}
  \caption{The attention maps of the Baseline trained on clean RAFDB,  our proposed framework trained on 30\% and our proposed framework trained on 60\% synthetic label noise of RAFDB on test images from RAFDB using Grad-CAM are compared in this figure. The emotion label in red color mean the prediction is incorrect and emotion label in green mean the prediction is correct. On top of every image, we have given the probability with which the model predicted the label.}
  \label{fig:gradcam_mutex}
\end{figure}

Attention Maps are the weighted feature maps,such that more weight is given to the salient regions which are more concentrated on by the model to predict the label. In order to investigate these salient regions focused by the proposed model, the attention-weighted activation maps are visualized using Grad-CAM \cite{grad} for Baseline trained on RAFDB and proposed method trained on 30\% and 60\% synthetic label noise on RAFDB. Darker color indicates high attention while lighter color indicates negligible attention.  The baseline sometimes focuses on irrelevant parts or misses out on relevant parts. In comparison to Baseline, the proposed model attends to non-occluded and relevant parts for expression recognition. These visualizations validate the effectiveness of our framework in the prediction of the correct label instead of over-fitting to the noisy label. In Fig. \ref {fig:gradcam_mutex}, the emotion given below  each image is the label that is predicted by model and green represents correct whereas rad represents incorrect prediction. On top of the every image, we have given the prediction probability with which, the model predicted the label. Clearly, Our method is able to learn robust features in the presence of noisy labels.
% Here, clean label refers to the actual label in the training set before changing it to the noisy label for training.

\begin{figure}[hbt!]
\centering
  \includegraphics[scale=0.5]{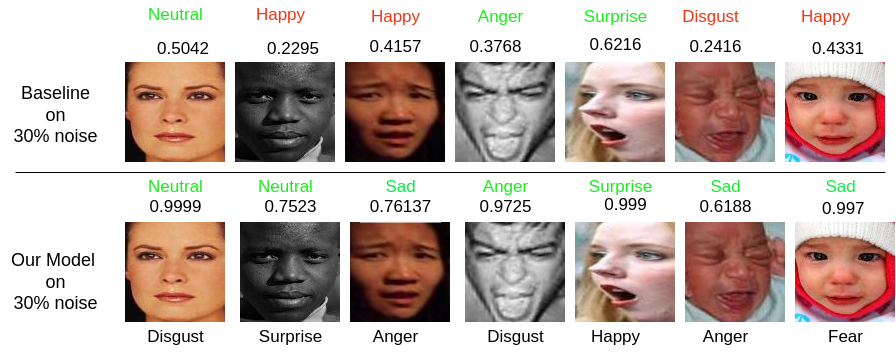}
  \caption{Prediction scores compared to baseline trained on 30\% label noise dataset on RAFDB and our model trained on 30\% label noise dataset on RAFDB. Above every image, we have given the prediction of the model mentioned on first column of each row. The emotion label in green represents correct and in red represents incorrect label. Just below the label, we have given the probability with which the model predicted the label. Below the down set of images, we have given the noisy label with which the image was trained. Inspite of giving different incorrect label, model is able to learn the correct label.}
  \label{fig:predscores_mutex}
\end{figure}

\subsubsection{Confidence Scores}
In order to quantitatively demonstrate the effectiveness of our model  with noisy labeled images, we visualize the prediction/confidence scores on images of different expression classes from the RAFDB dataset. These are shown in Fig. \ref{fig:predscores_mutex}. The more uncertain the annotation of a sample is, the lower will be its confidence score. Given any noisy label, our model DNFER predicts correctly the true label with high probability for almost all cases.

\section{Conclusions}\label{sec:conclusion}
In this paper, we have proposed a new method to handle the problem of noisy annotations in FER datasets. Our model uses posterior prediction probabilities from a positive class classifier and uses a dynamic adaptive threshold to get confident and non-confident samples. On the confident samples, we use CE loss, and on the non-confident samples using the prediction probabilities from the negative class classifier, we use consistency loss in each mini-match. Recent sample selection algorithms are computationally expensive since they employ many networks for joint or peer training and/or require knowledge of the noise rate beforehand. In contrast to previous models, our model doesn't need to know the noise rates nor needs to learn using multiple networks, nor need a separate supplement of clean data. Our model uses all the samples either for consistency loss on the predictions of the negative class classifier or supervised loss on the predictions of the positive class classifier. By using the dynamic adaptive threshold, It also handles the class imbalance problem, It also caters to inter-class similarities and intra-class difficulties. Instead of imposing the non-confident samples to learn the positive class which might be wrong in our case due to noisy labels, we impose consistency only on the negative classes of these non-confident samples but not on the positive class. To summarize, Our method on lower noise rates performs on par with the model that uses only confident samples based on the dynamic adaptive threshold one RAFDB but performs better in the case of FERPlus but it is more effective as the noise rate increases in the dataset as we can see that the improvement over the former by 0.433 to 4.537\% on FERPlus and 2.56 to 11.71\% on RAFDB datasets.

\section*{Acknowledgments}
We dedicate this work to Bhagawan Sri Sathya Sai Baba, Divine Founder Chancellor of Sri Sathya Sai Institute of Higher Learning, Prasanthi Nilayam, A.P., India.

%Bibliography
\bibliographystyle{unsrt}  
\bibliography{references}

\end{document}